\documentclass[11pt]{article}
\usepackage{geometry}               
\usepackage{amsmath} 
\usepackage{amssymb}  
\usepackage{mathrsfs}
\usepackage{mathtools}
\usepackage{xcolor}
\usepackage{algorithm}
\usepackage{algorithmic}
\usepackage{bm}
\usepackage{multirow}
\usepackage{booktabs}
\usepackage{url}

\usepackage{amsthm}

\newtheorem{theorem}{Theorem}
\newtheorem{proposition}{Proposition}

\newcommand{\obs}{o}

\let\vec\mathbf
\title{\LARGE \bf
Egocentric Conformal Prediction for Safe and Efficient Navigation in Dynamic Cluttered Environments\thanks{This work was supported in part by the Information and Communications Technology Planning and Evaluation
(IITP) grants funded by MSIT No. 2022-0-00124, No. 2022-0-00480 and No. RS-2021-II211343,
Artificial Intelligence Graduate School Program (Seoul National University).}
}

\author{Jaeuk Shin \and Jungjin Lee \and Insoon Yang
\thanks{The authors are with the Department of Electrical and Computer Engineering, ASRI,  Seoul National University, Seoul 08826, South Korea, 
        {\tt\small \{sju5379, jungbbal, insoonyang\}@snu.ac.kr}}
}

\date{}

\begin{document}

\maketitle
\thispagestyle{empty}
\pagestyle{empty}

 \begin{abstract}
Conformal prediction (CP) has emerged as a powerful tool in robotics and control, thanks to its ability to calibrate complex, data-driven models with formal guarantees. However, in robot navigation tasks, existing CP-based methods often decouple prediction from control, evaluating models without considering whether prediction errors actually compromise safety. Consequently, ego-vehicles may become overly conservative or even immobilized when all potential trajectories appear infeasible. To address this issue, we propose a novel CP-based navigation framework that responds exclusively to safety-critical prediction errors. Our approach introduces egocentric score functions that quantify how much closer obstacles are to a candidate vehicle position than anticipated. These score functions are then integrated into a model predictive control scheme, wherein each candidate state is individually evaluated for safety. Combined with an adaptive CP mechanism, our framework dynamically adjusts to changes in obstacle motion without resorting to unnecessary conservatism. Theoretical analyses indicate that our method outperforms existing CP-based approaches in terms of cost-efficiency while maintaining the desired safety levels, as further validated through experiments on real-world datasets featuring densely populated pedestrian environments.
\end{abstract}

\section{Introduction}

Data-driven predictive models are now an essential component for achieving safe autonomy in navigation, particularly given the complexity and uncertainty of dynamic environments.
Since safe control of ego-vehicles depends on accurately predicting the future states of surrounding dynamic agents, numerous motion forecasting models~\cite{salzmann2020trajectron++,yao2021bitrap} have been developed to forecast an agent’s future motions from historical data.
Nevertheless, these predictions remain inherently prone to error, primarily because they lack information about hidden contexts or intents---such as agents’ goals, velocity preferences, or even social relationships among human agents.

To address these limitations, \emph{conformal prediction} (CP)~\cite{vovk2005algorithmic,angelopoulos2021gentle} has been employed to reliably assess the models' predictive capabilities. 
The method offers a principled yet straightforward procedure for calibrating the models. At test time, the calibration results can be used to construct a confidence set that contains the true future states of the environment, assuming that the test and calibration data are exchangeable (i.e., their joint distribution is symmetric).
Consequently, CP has been successfully applied to a variety of problems, including reinforcement learning~\cite{strawn2023conformal,huang2024conformal}, linear systems~\cite{vlahakis2024conformal}, multi-agent systems~\cite{muthali2023multi}, and even the verification of extremely complex robotic planners based on large language models~\cite{ren2023robots,wang2023conformal}. Moreover, CP is particularly valuable for navigation applications, as it provides theoretical guarantees regarding system safety~\cite{lindemann2023safe,dixit2023adaptive,stamouli2024recursively}.

Unfortunately, the exchangeability assumption underpinning CP is frequently violated in real-world scenarios, mainly due to discrepancies between the environments in which the model is calibrated and those encountered at test time---for example, temporal changes in the motion patterns of dynamic obstacles or interactions between the ego-vehicle and obstacles. 
As a result, the environment often undergoes distributional shifts, and the ego-vehicle operating within it frequently encounters \emph{out-of-distribution} scenarios in which the confidence sets built from the calibration data are no longer valid.
To mitigate this issue, various strategies have been proposed that enable the ego-vehicle to recognize heterogeneous scenarios and adapt its behavior accordingly~\cite{sharma2023pac,contreras2024out,huang2024conformal,lekeufack2024conformal}.
Among these approaches, \emph{adaptive conformal prediction} (ACP) methods~\cite{gibbs2021adaptive} allow the CP procedure to adjust to observed temporal shifts. These methods are flexible enough to accommodate various types of distributional drift while maintaining asymptotic coverage guarantees, and have been used in the control domain~\cite{dixit2023adaptive}.
Despite these advantages, control schemes based on ACP methods are known to yield large confidence sets when distribution  shifts occur, which may degrade vehicle efficiency~\cite{lekeufack2024conformal} or even cause the vehicle to freeze.
This limitation has spurred extensive research into enhancing ACP's adaptation speed, either by leveraging robust online learning techniques~\cite{zaffran2022adaptive,bhatnagar2023improved,angelopoulos2023conformal,gibbs2024conformal} or by designing more effective score functions to obtain tighter confidence sets~\cite{stamouli2024recursively,zhou2024conformalized}.

Addressing the conservativeness of ACP-based control methods, we adopt an alternative perspective focused on how CP is integrated into motion control. A primary shortcoming of existing CP-based approaches is that prediction models are calibrated in an \emph{obstacle-centric} manner; that is, CP procedures typically rely on score functions that measure the discrepancy between predicted and true states. Consequently, any large prediction error results in a reduction of the feasible vehicle state space---even when such errors do not lead to hazardous planning decisions. This effect is exacerbated during abrupt distributional changes, as illustrated in Figure~\ref{fig:feasible-sets}.

To overcome this limitation, we propose a novel CP method, termed \emph{egocentric conformal prediction} (ECP), which adopts an egocentric approach to model calibration. We introduce a  {state-dependent} score function that captures only the safety-related error by quantifying how much closer dynamic obstacles are to a given state than predicted by the model. By design, this score function prevents cost-efficient motions from being excluded due to safety-irrelevant prediction errors. Moreover, since each vehicle state is associated with a unique score function, the safety of multiple states can be evaluated concurrently. As a result, the score function defines a map over the state space that can be employed by a safety-aware controller---such as {model predictive control (MPC)}---to assess the feasibility of each state. Unlike existing risk maps for robot navigation~\cite{hakobyan2023, allen2023}, our map is \emph{conformalized} and, when combined with ACP, can efficiently track changes in environmental distribution. Despite its simplicity, the proposed score function consistently yields larger safe sets compared to traditional approaches, thereby enabling the vehicle to adopt more aggressive control inputs. A recently proposed online optimization method, known as the conformal controller~\cite{lekeufack2024conformal}, shares a similar philosophy by directly calibrating the risk associated with controller decisions; however, while that method relies on an unconstrained MPC formulation, our approach is built upon a constrained MPC framework. By avoiding a penalized MPC formulation---which necessitates careful tuning of weight parameters---our method effectively balances efficiency and safety without excessive parameter tuning.

To leverage ECP for control, we introduce a novel MPC method, termed ECP-MPC. In practical implementations, directly computing the conformalized map may require a fine discretization of the state space, which becomes computationally intractable in large-scale environments. To address this, we propose a tractable approach that discretizes the input space instead of the state space. The resulting ECP-MPC operates with sufficient speed for deployment in real-world scenarios. Moreover, we show that our method is more cost-efficient than the original ACP-MPC while maintaining an asymptotic safety guarantee. Finally, we demonstrate the empirical effectiveness of our strategy in highly cluttered navigation scenarios.

The remainder of the paper is organized as follows. In Section~\ref{sec:problem-setting}, we present a mathematical formulation of the navigation problem in dynamic environments. Section~\ref{sec:cp-mpc} describes an MPC method based on CP and its adaptive variant, along with a discussion of their limitations. In Section~\ref{sec:ecp-mpc}, we introduce the egocentric model calibration procedure and the associated MPC problem, and propose a tractable approximation to mitigate the computational burden. Finally, Section~\ref{sec:exp-res} empirically evaluates the cost-efficiency and safety of our proposed method in crowded scenarios.

\section{The Setup}\label{sec:problem-setting}

\subsection{Notations}

The notation $\mathbb{I}[\cdot]$ denotes indicator functions, and $\delta_{s}$ represents the Dirac measure concentrated at $s \in \mathbb{R}$.
Given a probability distribution $P$ over $\mathbb{R}$ and $\alpha \in (0,1)$, we define the $\alpha$-quantile of $P$ as
$\mathsf{Quantile}_{\alpha}(P) \coloneqq \inf\{s\in\mathbb{R}: P(-\infty, s] \geq \alpha \}$.
Abusing notation slightly, we also write $\mathsf{Quantile}_{\alpha}(\mathscr{S})$ for a set $\mathscr{S} \subseteq \mathbb{R}$ to denote the $\alpha$-quantile of the empirical distribution $1/|\mathscr{S}|\sum_{s\in \mathscr{S}}\delta_s$. 
For $\vec{x} \in \mathbb{R}^d$ and $\vec{Y} \subseteq \mathbb{R}^d$, the minimum distance between $\vec{x}$ and $\vec{Y}$ is denoted as $d(\vec{x}, \vec{Y}) = \inf_{\vec{y} \in \vec{Y}} \|\vec{x} - \vec{y} \|_2$. 
The Hausdorff distance between sets $\vec{Y}, \vec{Y}' \subseteq \mathbb{R}^d$ is defined as
$d_H (\vec{Y}, \vec{Y}') = \min\{ \sup_{\vec{y}\in \vec{Y}} d(\vec{y}, \vec{Y}'), \sup_{\vec{y}'\in \vec{Y}'} d(\vec{y}', \vec{Y})  \}$.

\begin{figure*}[t]
    \centering
    \includegraphics[width=0.94\linewidth]{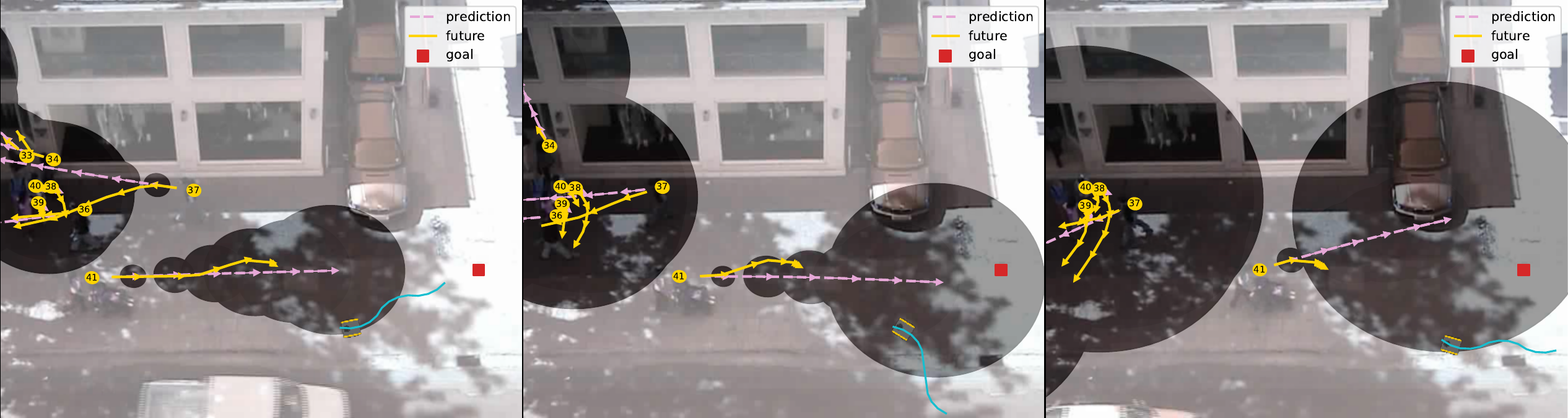}
    \caption{Example of ACP-MPC applied to a mobile navigation scenario built from the UCY \texttt{zara2} dataset~\cite{lerner2007crowds}, shown in temporal order. The predicted trajectories, denoted as $\vec{Y}_{t+i|t}$, are drawn as pink dashed lines, while the corresponding ground truth positions, $\vec{Y}_{t+i}$, are depicted as yellow solid lines. The shaded circles represent the confidence sets $\mathsf{C}^{\text{obs}}_{t+i|t}$ generated by ACP. Even though the observed prediction errors are not safety-critical, they cause the ACP parameters to decrease and the confidence sets to expand over time, leading to increasingly conservative vehicle motion plans (in cyan).}
    \label{fig:acp-mpc-example}
\end{figure*}

\subsection{Navigation in Dynamic Environments} 

We consider the problem of controlling an ego-vehicle to reach a specified goal in the presence of multiple dynamic obstacles. We denote the state space of the ego-vehicle by $\mathcal{X}$ and the observation space of the moving obstacles by $\mathcal{Y}$. For simplicity, we assume $\mathcal{X} = \mathbb{R}^2$, meaning that the robot dynamics are completely characterized by its 2D position. Thus, $\vec{x}_t \in \mathcal{X}$ represents the position of the robot at time $t$. The discrete-time dynamics of the robot is given by
\[
\vec{x}_{t+1} = f_r(\vec{x}_t, \vec{u}_t), \quad t = 0, 1, \ldots,
\]
where $\vec{u}_t$ is chosen from the input space $\mathcal{U}$.

We assume that the perception system is perfect; that is, the 2D position of each moving obstacle in the scene is detected and tracked without error. Denote the finite set of all moving obstacle IDs by $\mathcal{O}$. For each $\obs \in \mathcal{O}$, the measured position of moving obstacle $\obs$ at time $t$ is given by $\vec{y}^{\obs}_t \in \mathbb{R}^2$. The overall observation of the moving obstacles is then defined as
\[
\vec{Y}_t \coloneqq \left\{\vec{y}^{\obs}_t : \obs \in \mathcal{O}\right\},
\]
and we let $\mathcal{Y} = (\mathbb{R}^2)^{\mathcal{O}}$. In practice, each moving obstacle can be correctly labeled by applying fine-grained tracking methods to the detected crowd.

To account for the variability in moving obstacle motions, we model the sequence $\{\vec{Y}_t\}_{t\geq 0}$ as a random process without imposing restrictive assumptions, such as the Markov property. To ensure the safety of the system, the robot is subject to the \emph{collision-avoidance} constraints
\[
\vec{x}_t \in \mathcal{S}_{t} \coloneqq \left\{\vec{x} \in \mathcal{X} : \|\vec{x} - \vec{y}^{\obs}_t\| \geq r_{\text{safe}} \quad \forall\, \obs \in \mathcal{O} \right\},
\]
where $r_{\text{safe}} > 0$ is a user-specified safety margin.

We assume access to a deterministic prediction model $\hat{f}: \mathcal{Y}^H \to \mathcal{Y}^N$ that takes as input the moving obstacles' motion history of length $H$ and outputs predictions for the next $N$ steps. Throughout the paper, we fix a single prediction model $\hat{f}$ and adopt the notation
\[
\left(\vec{Y}_{t+1|t}, \ldots, \vec{Y}_{t+N|t}\right) = \hat{f}(\vec{h}_t) = \hat{f}\left(\vec{Y}_{t-H+1}, \ldots, \vec{Y}_t\right).
\]
That is, $\vec{Y}_{t+i|t} = \hat{f}_i(\vec{h}_t)$ denotes the $i$-th step prediction given the history $\vec{h}_t \coloneqq (\vec{Y}_{t-H+1}, \ldots, \vec{Y}_t)$.

\section{CP for Navigation in Dynamic Environments}\label{sec:cp-mpc}

\subsection{Brief Review of CP-MPC}

Since the predictions $\vec{Y}_{t+i|t}$ generated by $\hat{f}$ often differ from the true values $\vec{Y}_{t+i}$, a naive use of these predictions may render the system unsafe. 
To address this, CP is employed to systematically compensate for potential prediction errors by constructing a confidence region that contains $\vec{Y}_{t+i}$ with probability $1-\alpha$, where $\alpha \in (0, 1)$ is a user-specified miscoverage level~\cite{lindemann2023safe,dixit2023adaptive}. 
In particular, a safe-MPC framework that integrates CP was proposed in~\cite{lindemann2023safe}, which we refer to as  {CP-MPC}.
The method employs the following  {score function} to quantify the $i$-th step prediction error of $\hat{f}$:
\begin{equation}\label{eq:original-score}
s_i(\vec{h}, \vec{Y}) \coloneqq \max_{\obs\in \mathcal{O}} \big\| \vec{y}^{\obs} - \hat{f}_i(\vec{h})^{\obs}\big\|, \quad \vec{h} \in \mathcal{Y}^H, \quad \vec{Y} \in \mathcal{Y}.
\end{equation}

Assuming that an offline dataset of obstacle trajectories $\{\vec{Y}^{(k)}_t\}_{k=1}^n$ from $n$ different scenarios is available, we construct a calibration set 
\[
\mathscr{D}^{\text{offline}}_{t+i|t} \coloneqq  \big \{ (\vec{h}^{(k)}_{t}, \vec{Y}^{(k)}_{t+i}) \big \}_{k=1}^n
\]
to evaluate the quality of $\hat{f}_i$ for each $t \geq 0$ and $1 \leq i \leq N$.
The calibration of $\hat{f}_i$ is performed as follows: First, we compute the set of scores
\[
\mathscr{S}_{t+i|t} = \big\{ s_i (\vec{h}^{(k)}_{t}, \vec{Y}^{(k)}_{t+i})\big\}_{k=1}^n,
\]
and then estimate the $(\lceil (n+1)(1-\alpha) \rceil/n)$-quantile of this score distribution:
\[
\mathsf{R}_{t+i|t} \coloneqq \mathsf{Quantile}_{\big(\lceil (n+1)(1-\alpha) \rceil/n\big)}\left(\mathscr{S}_{t+i|t}\right).
\]
When a test input $\vec{h}_t$ is observed, we construct the confidence set
\[
\mathsf{C}_{t+i|t} \coloneqq\big \{ \vec{Y} \in \mathcal{Y} : s_i(\vec{h}_{t}, \vec{Y}) \leq \mathsf{R}_{t+i|t} \big\}
\]
that covers $\vec{Y}_{t+i}$ with high probability. 
Indeed, under the assumption that 
\[
\mathscr{D}_{t+i|t} \cup \{ (\vec{h}_t, \vec{Y}_{t+i})\}
\]
is independent and identically distributed (i.i.d.) or, at least, exchangeable, the following marginal inequality holds:
\begin{align}\label{eq:cp-mpc-coverage}
\mathbb{P}\Big[ \vec{Y}_{t+i} \in \mathsf{C}_{t+i|t} \Big] &= \mathbb{P}\Big[ \max_{\obs\in \mathcal{O}} \big\| \vec{y}^{\obs}_{t+i|t} - \vec{y}^{\obs}_{t+i} \big\| \leq \mathsf{R}_{t+i|t} \Big] \notag \\
&\geq 1 - \alpha,
\end{align}
where $\mathbb{P}$ denotes the randomness over both $\mathscr{D}^{\text{offline}}_{t+i|t}$ and $(\vec{h}_t, \vec{Y}_{t+i})$.

Based on this procedure, a shrinking-horizon MPC was proposed in~\cite{lindemann2023safe} to solve finite-horizon navigation problems\footnote{In~\cite{stamouli2024recursively}, the score~\eqref{eq:original-score} was modified to ensure the recursive feasibility of~\eqref{eq:cp-mpc}.}:
\begin{align}\label{eq:cp-mpc}
J^{\text{obs}}(\vec{x}_t) \coloneqq \min_{\vec{x}, \vec{u}} & \sum_{i=0}^{T-t-1} \ell\big(\vec{x}_{t+i|t}, \vec{u}_{t+i|t}\big) + \ell_f\big(\vec{x}_{T|t}\big) \notag \\
\text{s.t.} \quad & \vec{x}_{t|t} = \vec{x}_t, \notag \\
& \vec{x}_{t+i+1|t} = f_r\big(\vec{x}_{t+i|t}, \vec{u}_{t+i|t}\big), \quad 0 \leq i < T-t, \notag \\
& \vec{u}_{t+i|t} \in \mathcal{U}, \quad 0 \leq i < T-t, \notag \\
& \vec{x}_{t+i|t} \in \mathcal{S}^{\text{obs}}_{t+i|t}, \quad 1 \leq i < T-t.
\end{align}
Here, $T > 0$ denotes the horizon length, $\ell: \mathcal{X} \times \mathcal{U} \to \mathbb{R}$ is an intermediate cost function, and $\ell_f: \mathcal{X} \to \mathbb{R}$ is a terminal cost function. 
The set $\mathcal{S}^{\text{obs}}_{t+i|t}$ represents the safe set that incorporates the calibration information $\mathsf{R}_{t+i|t}$:
\begin{equation}\label{eq:obs-cp-safe-set}
\mathcal{S}^{\text{obs}}_{t+i|t} \coloneqq \big \{ \vec{x} \in \mathcal{X} : d\big(\vec{x}_{t+i|t}, \vec{Y}_{t+i|t}\big) \geq r_{\text{safe}} + \mathsf{R}_{t+i|t} \big \}.
\end{equation}
We use the superscript $\text{obs}$ (short for ``obstacle") to distinguish~\eqref{eq:obs-cp-safe-set} from the safe set defined later. Assuming the feasibility of~\eqref{eq:cp-mpc} for all $0 \leq t < T$ and using~\eqref{eq:cp-mpc-coverage},~\cite{lindemann2023safe} proved that the corresponding closed-loop system is safe:
\[
\mathbb{P}\Big[ \vec{x}_t \in \mathcal{S}_t \; \forall\, t \in [0, T] \Big] \geq 1 - T\cdot\alpha.
\]

However, the i.i.d. (or exchangeability) assumption may be too restrictive in realistic scenarios, degrading safety as the constructed confidence sets may no longer be valid~\cite{lindemann2023safe,dixit2023adaptive}. 
Although such assumptions hold when the calibration set and test data are statistically equivalent, the ego-vehicle is likely to encounter different obstacle motion patterns---such as when interactions occur between the ego-vehicle and obstacles, or when obstacles behave drastically differently from those in the offline dataset. 
To address this issue,~\cite{dixit2023adaptive} integrated ACP into CP-MPC.
The method modifies~\cite{lindemann2023safe} in two key aspects: First, the score function~\eqref{eq:original-score} is evaluated on an online dataset 
\[
\mathscr{D}_{t+i|t} \coloneqq \{ (\vec{h}_{t'-i}, \vec{Y}_{t'}) \}_{t' = t - M + 1}^{t}
\]
of size $M$ at time $t$, instead of using $\mathscr{D}^{\text{offline}}_{t+i|t}$.
This adjustment accounts for temporal shifts in obstacle motions that may not be captured by the offline dataset.
Second, the $(1-\alpha^i_t)$-quantile of the set 
\[
\mathscr{S}_{t+i|t} \coloneqq \{ s_i(\vec{h}_{t'-i}, \vec{Y}_{t'}) \}_{t' = t - M + 1}^{t}
\]
is computed, i.e., 
\[
\mathsf{R}_{t+i|t} \coloneqq \mathsf{Quantile}_{1-\alpha^i_t}\big(\mathscr{S}_{t+i|t}\big),
\]
where the parameters $\alpha^i_t$ are updated online to ensure asymptotic $(1-\alpha)$-coverage. 
Specifically, each $\alpha^i_t$ is updated by incorporating a new observation $\vec{Y}_t$ and checking whether the previously proposed confidence set
\[
\mathsf{C}^{\text{obs}}_{t|t-i} \coloneqq \{ \vec{Y} \in \mathcal{Y} : s_i(\vec{h}_{t-i}, \vec{Y}) \leq \mathsf{R}_{t|t-i} \}
\]
is valid:
\[
\alpha^{i}_{t+1}(\vec{x}) = \alpha^{i}_{t}(\vec{x}) + \gamma \Big( \alpha - \mathbb{I}\big[ \vec{Y}_{t} \notin \mathsf{C}^{\text{obs}}_{t|t-i} \big] \Big),
\]
where $\gamma > 0$ is a step size. 
The resulting confidence sets $\mathsf{C}^{\text{obs}}_{t+i|t}$ achieve the following {asymptotic} coverage guarantee for all $1 \leq i \leq N$ with probability 1~\cite{dixit2023adaptive}:
\begin{equation}\label{eq:original-cp-asymptotic}
\lim_{t \to\infty} \frac{1}{T} \sum_{t=1}^T \mathbb{I}\Big[ \vec{Y}_{t+i} \in \mathsf{C}^{\text{obs}}_{t+i|t} \Big] = 1 - \alpha.
\end{equation}

This motivates the following receding-horizon MPC formulation, referred to as {ACP-MPC} throughout this paper (which is now applicable to infinite-horizon problems):
\begin{align}\label{eq:acp-mpc}
J^{\text{obs}}_t(\vec{x}_t) \coloneqq \min_{\vec{x}, \vec{u}} \ & \sum_{i=0}^{N} \ell\big(\vec{x}_{t+i|t}, \vec{u}_{t+i|t}\big) + \ell_f\big(\vec{x}_{t+N|t}\big) \notag \\
\text{s.t.} \quad & \vec{x}_{t|t} = \vec{x}_t, \notag \\
& \vec{x}_{t+i+1|t} = f_r\big(\vec{x}_{t+i|t}, \vec{u}_{t+i|t}\big), \quad 0 \leq i \leq N - 1, \notag \\
& \vec{u}_{t+i|t} \in \mathcal{U}, \quad 0 \leq i \leq N - 1, \notag \\
& \vec{x}_{t+i|t} \in \mathcal{S}^{\text{obs}}_{t+i|t}, \quad 1 \leq i \leq N.
\end{align}
Here, the safety constraint $\mathcal{S}^{\text{obs}}_{t+i|t}$ is defined as in \eqref{eq:obs-cp-safe-set}, with $\mathsf{R}_{t+i|t}$ computed via the online procedure described above.

As shown in~\cite{dixit2023adaptive}, the safety constraints in \eqref{eq:acp-mpc} together with the asymptotic guarantee \eqref{eq:original-cp-asymptotic} ensure that, assuming the feasibility of \eqref{eq:acp-mpc} for all $t \geq 0$, the closed-loop system satisfies
\[
\liminf_{t\to\infty} \frac{1}{T} \sum_{t=0}^{T-1} \mathbb{I} \Big[ \vec{x}_t \in \mathcal{S}_t \Big] \geq 1 - \alpha, \quad \text{w.p. 1}.
\]

\subsection{Conservativeness of ACP-MPC}

Unfortunately, the controller~\eqref{eq:acp-mpc} has been observed to yield overly conservative behavior.
Fig.~\ref{fig:acp-mpc-example} illustrates a scenario in which the rapid expansion of the confidence sets obtained through CP forces the ego-vehicle to plan highly inefficient paths, corroborating the findings in~\cite{lekeufack2024conformal}. Although multiple factors may contribute to this conservativeness, one key limitation arises from calibrating model error in a purely obstacle-centric manner. For instance, suppose that an obstacle is receding from the robot, yet the model erroneously predicts that the obstacle is approaching. Although following this incorrect prediction and choosing an avoidant action does not compromise safety, the $\ell^2$-norm error is still incorporated into the score, causing the confidence set to inflate. Consequently, the vehicle’s feasible control inputs become overly restricted.

Another potential factor specific to the definition~\eqref{eq:original-score} is that prediction errors for obstacles far from the robot and those nearby contribute equally to the size of the confidence sets, even though errors in the positions of distant obstacles are typically less critical. This suggests that model errors should be evaluated in an egocentric fashion, so that the confidence set expands primarily in response to genuinely safety-critical errors.

\section{Egocentric Conformal Prediction for Safe and Efficient Navigation}\label{sec:ecp-mpc}
\subsection{Egocentric Model Calibration}

\begin{figure}[t!]
    \centering
    \includegraphics[width=0.85\linewidth]{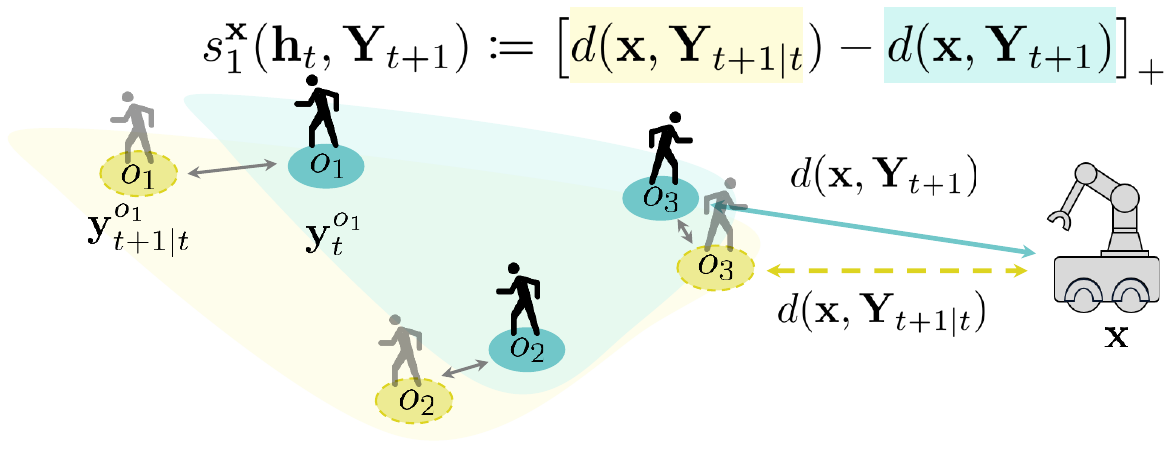}
    \caption{Illustration of the egocentric score function in a scenario with three moving obstacles. In our formulation, errors for obstacles $v_1$ and $v_2$ are neglected since they are far from a given state $\vec{x}$, making these errors irrelevant for assessing its safety. Although a prediction error occurs for $v_3$, the obstacle closest to $\vec{x}$, it does not affect our score function because the true future position is farther from $\vec{x}$ than the predicted position. Consequently, $s^{\vec{x}}_1 = 0$ in this case.}
    \label{fig:score-function}
\end{figure}

To address the conservativeness issue in CP-MPC, we introduce \emph{egocentric conformal prediction} (ECP), which applies ACP to errors measured relative to candidate states of the ego-vehicle. The key idea is to design a distinct score function for each candidate position and focus on the minimum distance between the robot and the moving obstacles. Specifically, we define an {egocentric score function associated with} $\vec{x} \in \mathcal{X}$ as
\begin{equation}\label{eq:egocentric-score}
s^{\vec{x}}_i(\vec{h}, \vec{Y}) \coloneqq \Bigl[ d\big(\vec{x}, \hat{f}_i(\vec{h})\big) - d\big(\vec{x}, \vec{Y}\big) \Bigr]_+,
\end{equation}
where $\vec{h} \in \mathcal{Y}^H$, $\vec{Y} \in \mathcal{Y}$, and $[x]_+ = \max(x, 0)$. Figure~\ref{fig:score-function} provides a sketch of our score function. Given an observed history $\vec{h}_t$ up to time $t$ and the true future obstacle positions $\vec{Y}_{t+i}$ (realized at time $t+i$), the score
\[
s^{\vec{x}}_i(\vec{h}_t, \vec{Y}_{t+i}) = \Bigl[ d\big(\vec{x}, \vec{Y}_{t+i|t}\big) - d\big(\vec{x}, \vec{Y}_{t+i}\big) \Bigr]_+
\]
quantifies the extent to which, when located at $\vec{x}$, the robot is closer to the obstacles than anticipated. In contrast to the single score function defined in~\eqref{eq:original-score}, we now have a family of score functions $\{ s^{\vec{x}}_i \}_{\vec{x} \in \mathcal{X}}$ for each $i = 1,\ldots,N$. Each score function $s^{\vec{x}}_i$ penalizes the case where the predicted minimum distance between the robot and the obstacles exceeds the true distance, potentially leading to unsafe control inputs. Notably, our new formulation only considers prediction errors relative to the state of the vehicle, thereby avoiding penalization for \emph{safety-irrelevant} errors.

Based on the egocentric score functions~\eqref{eq:egocentric-score}, we run the ACP calibration procedure to compute
\[
\mathsf{R}^{\vec{x}}_{t+i|t} \coloneqq \mathsf{Quantile}_{1-\alpha^i_t(\vec{x})}\Bigl( \mathscr{S}^{\vec{x}}_{t+i|t} \Bigr),
\]
for each $\vec{x} \in \mathcal{X}$, where
\[
\mathscr{S}^{\vec{x}}_{t+i|t} \coloneqq \Bigl\{ s^{\vec{x}}_i\big(\vec{h}_{t'-i}, \vec{Y}_{t'}\big) \Bigr\}_{t' = t - M + 1}^{t}.
\]
The ACP parameters $\alpha^i_t(\vec{x})$ are now defined for each $\vec{x} \in \mathcal{X}$ and are updated for all $1 \leq i \leq N$ as follows:
\[
\alpha^{i}_{t+1}(\vec{x}) = \alpha^{i}_{t}(\vec{x}) + \gamma \Bigl( \alpha - \mathbb{I}\Bigl[ \vec{Y}_{t} \notin \mathsf{C}^{\vec{x}}_{t|t-i} \Bigr] \Bigr), \quad t \geq 0,
\]
where $\gamma > 0$ is a step size, and the confidence set is defined as
\[
\mathsf{C}^{\vec{x}}_{t+i|t} \coloneqq \Bigl\{ \vec{Y} \in \mathcal{Y} : s^{\vec{x}}_i(\vec{h}_{t}, \vec{Y}) \leq \mathsf{R}^{\vec{x}}_{t+i|t} \Bigr\}.
\]

\begin{figure}[t]
\centering
\includegraphics[width=0.9\textwidth]{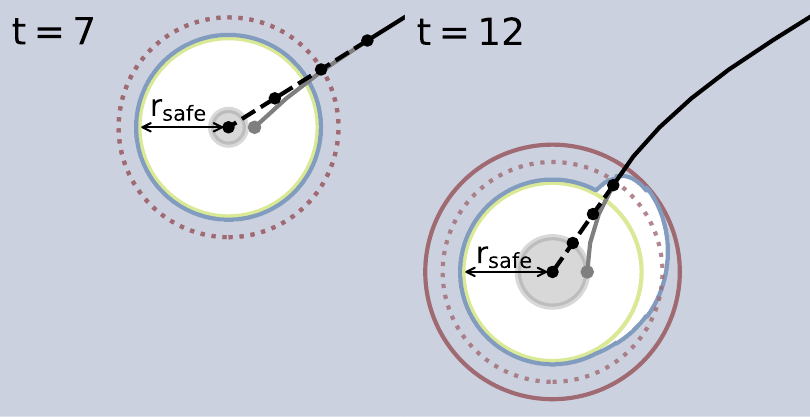}
\includegraphics[width=0.9\textwidth]{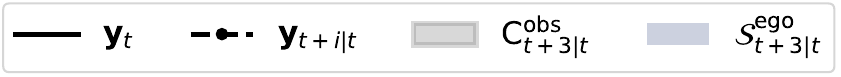}
\caption{Comparison between the obstacle-centric safety constraint $\mathcal{S}^{\text{obs}}_{t+i|t}$ and its egocentric counterpart $\mathcal{S}^{\text{ego}}_{t+i|t}$. The gray area represents the obstacle-centric confidence set $\mathsf{C}^{\text{obs}}_{t+i|t}$ proposed by the original ACP (with its boundary shown in red). In this scenario, an obstacle undergoes abrupt velocity changes, causing $\mathsf{C}^{\text{obs}}_{t+3|t}$ to fail to cover the true future position $\vec{y}_{t+3}$ for $t \in [7, 9]$. Consequently, the ACP parameter $\alpha^3_t$ diminishes during $t \in [10, 12]$, and the size of $\mathcal{S}^{\text{obs}}_{t+3|t}$ rapidly shrinks from the region outlined by the red dotted boundary to that with the red solid boundary. In contrast, the egocentric score~\eqref{eq:egocentric-score} increases only for $\vec{x}$ on the right of the obstacle, since the prediction error is harmless for states on the left side. As a result, $\mathcal{S}^{\text{ego}}_{t+3|t}$ shrinks only from the right side.}
\label{fig:feasible-sets}
\end{figure}

This procedure leads to the following safety constraint:
\begin{equation}\label{eq:egocentric-safe-set}
\mathcal{S}^{\text{ego}}_{t+i|t} \coloneqq \Bigl\{ \vec{x} \in \mathcal{X} : d\big(\vec{x}, \vec{Y}_{t+i|t}\big) \geq r_{\text{safe}} + \mathsf{R}^{\vec{x}}_{t+i|t} \Bigr\}.
\end{equation}
This constraint yields the egocentric counterpart of~\eqref{eq:acp-mpc}:
\begin{align}\label{eq:ecp-mpc-naive}
J^{\text{ego}}(\vec{x}_t) \coloneqq \min_{\vec{x}, \vec{u}} \ & \sum_{i=0}^{N-1} \ell\big(\vec{x}_{t+i|t}, \vec{u}_{t+i|t}\big) + \ell_f\big(\vec{x}_{t+N|t}\big) \notag \\
\text{s.t.} \quad & \vec{x}_{t|t} = \vec{x}_t, \notag \\
& \vec{x}_{t+i+1|t} = f_r\big(\vec{x}_{t+i|t}, \vec{u}_{t+i|t}\big), \quad 0 \leq i \leq N-1, \notag \\
& \vec{x}_{t+i|t} \in \mathcal{X}, \quad 0 \leq i \leq N, \notag \\
& \vec{x}_{t+i|t} \in \mathcal{S}^{\text{ego}}_{t+i|t}, \quad 1 \leq i \leq N.
\end{align}
The key distinction between~\eqref{eq:ecp-mpc-naive} and the original CP-MPC formulation~\eqref{eq:acp-mpc} is in the shape of the safe set: the size of the confidence interval $\mathsf{R}^{\vec{x}}_{t+i|t}$ now depends on $\vec{x}$, so it can be viewed as a mapping from $\mathcal{X}$ to $\mathbb{R}$ that serves as a safety certificate for a given state. Augmented by the safety margin $r_{\text{safe}}$, this map determines the allowable distance between the vehicle and the dynamic obstacles, thereby defining the shape of the safe set. Figure~\ref{fig:feasible-sets} shows an example where $\mathcal{S}^{\text{ego}}_{t+i|t}$ is significantly larger than the obstacle-centric safe set $\mathcal{S}^{\text{obs}}_{t+i|t}$ while still responding appropriately to safety-critical errors. The following proposition confirms this observation, suggesting that the modified MPC~\eqref{eq:ecp-mpc-naive} is more efficient than the original CP-MPC.

\begin{proposition}\label{prop:less-conservative}
For any given history $\vec{h}_t$ and true future observation $\vec{Y}_{t+i}$, the egocentric score $s^{\vec{x}}_i(\vec{h}_t, \vec{Y}_{t+i})$ is bounded above by the Hausdorff distance between $\vec{Y}_{t+i}$ and $\vec{Y}_{t+i|t}$. In particular, we have
\begin{equation}\label{eq:score-comparison}
s^{\vec{x}}_i(\vec{h}_t, \vec{Y}_{t+i}) \leq d_H\big(\vec{Y}_{t+i}, \vec{Y}_{t+i|t}\big) \leq s_i(\vec{h}_t, \vec{Y}_{t+i}),
\end{equation}
where $s_i$ is defined in~\eqref{eq:original-score}. Hence, given the same calibration set $\mathscr{D}_{t+i|t}$, history $\vec{h}_t$, and parameters $\alpha^i_t(\vec{x}) \equiv \alpha^i_t$, the egocentric safe set $\mathcal{S}^{\text{ego}}_{t+i|t}$ is larger than or equal to the obstacle-centric safe set $\mathcal{S}^{\text{obs}}_{t+i|t}$, implying that
\[
J^{\text{ego}}(\vec{x}_t) \leq J^{\text{obs}}(\vec{x}_t).
\]
\end{proposition}

\begin{proof}
The inequality in~\eqref{eq:score-comparison} follows directly from the definitions in~\eqref{eq:original-score} and~\eqref{eq:egocentric-score}, together with the triangle inequality satisfied by the Hausdorff distance. Specifically, we have
\[
d\big(\vec{x}, \vec{Y}_{t+i|t}\big) - d\big(\vec{x}, \vec{Y}_{t+i}\big) \leq d_H\big(\vec{Y}_{t+i}, \vec{Y}_{t+i|t}\big),
\]
which implies the first inequality. Furthermore, for each $\obs \in \mathcal{O}$, 
\[
d\big(\vec{y}^{\obs}_{t+i}, \vec{Y}_{t+i|t}\big) \leq \|\vec{y}^{\obs}_{t+i} - \vec{y}^{\obs}_{t+i|t}\| \leq s_i(\vec{h}_t, \vec{Y}_{t+i}),
\]
so that
\[
d_H\big(\vec{Y}_{t+i}, \vec{Y}_{t+i|t}\big) \leq \max_{\vec{y} \in \vec{Y}_{t+i}} d\big(\vec{y}, \vec{Y}_{t+i|t}\big) \leq s_i(\vec{h}_t, \vec{Y}_{t+i}).
\]
Therefore, under the same calibration set $\mathscr{D}_{t+i|t}$ and the condition $\alpha^i_t(\vec{x}) \equiv \alpha^i_t$, we obtain
\[
\mathsf{R}^{\vec{x}}_{t+i|t} = \mathsf{Quantile}_{1-\alpha^i_t(\vec{x})}\Bigl(\mathscr{S}^{\text{ego}}_{t+i|t}\Bigr)
\leq \mathsf{Quantile}_{1-\alpha^i_t}\Bigl(\mathscr{S}_{t+i|t}\Bigr) = \mathsf{R}_{t+i|t}, \quad \forall\, \vec{x} \in \mathcal{X}.
\]
Hence, we conclude that
\[
\mathcal{S}^{\text{obs}}_{t+i|t} \subseteq \mathcal{S}^{\text{ego}}_{t+i|t},
\]
which implies that
\[
J^{\text{ego}}(\vec{x}_t) \leq J^{\text{obs}}(\vec{x}_t).
\]
\end{proof}

\subsection{ECP-MPC}

While the MPC problem~\eqref{eq:ecp-mpc-naive} may be preferred to~\eqref{eq:acp-mpc} in terms of cost efficiency, acquiring an exact solution to~\eqref{eq:ecp-mpc-naive} is challenging. Widely used gradient-based methods for solving such an MPC problem often require derivatives (or higher-order derivatives) of the functions defining the constraints---in our case, of $\mathsf{R}^{\vec{x}}_{t+i|t}$ with respect to $\vec{x}$. Although one can differentiate through $\mathsf{Quantile}$ and the score function $s^{\vec{x}}_i$ using automatic differentiation and soft-regularization techniques (e.g., as in~\cite{blondel2020fast, stutzlearning, sharma2023pac}), differentiating $\mathsf{R}^{\vec{x}}_{t+i|t}$ poses a subtle yet crucial challenge because of the dependency of $\alpha_t^i$ on $\vec{x}$. Since $\alpha_t^i(\vec{x})$ is time-varying, it must be updated for all $\vec{x}$ simultaneously, which often necessitates numerical differentiation methods. One feasible approach is to discretize $\mathcal{X}$, update $\alpha_t^i(\vec{x})$, and compute $\mathsf{R}^{\vec{x}}_{t+i|t}$ on this discrete set while adjusting $\mathcal{S}^{\text{ego}}_{t+i|t}$ to account for the resulting discretization error.\footnote{A high-level sketch of this approach is provided in Appendix~\ref{app:state-discretization}.} Unfortunately, this strategy is prohibitively expensive and practical only in small environments, where high-resolution discretization can mitigate the side effects.

\begin{algorithm}[t!]
\caption{ECP-MPC algorithm}
\label{alg:ecp-mpc}
\begin{algorithmic}[1]
	\STATE \textbf{Input}: history length $H$, prediction length $N$, prediction model $\hat{f}$;
    \STATE Collect initial observations $\vec{Y}_0, \ldots, \vec{Y}_{H+N-1}$;
	\FOR {$t = H + N, H+N+1, \ldots$}
              \STATE Observe $\vec{x}_t$ and $\vec{Y}_t$;
              \STATE Evaluate $\mathscr{S}^{\bm{\phi}}_{t+i|t}$~\eqref{eq:score-data} for all $1 \leq i \leq N$ and ${\bm{\phi}} \in \Phi$;
              \STATE Estimate the quantiles $\mathsf{R}^{\bm{\phi}}_{t+i|t}$~\eqref{eq:egocentric-quantile} for all $1 \leq i \leq N$ and ${\bm{\phi}} \in \Phi$;
              \STATE Update the ACP parameters:
              \[
              \alpha^{\bm{\phi}}_{t+i+1|t+1} = \alpha^{\bm{\phi}}_{t+i|t} + \gamma \Bigl( \alpha - \mathbb{I}\Bigl[ \vec{Y}_{t} \notin \mathsf{C}^{\bm{\phi}}_{t|t-i} \Bigr] \Bigr), \quad \forall\, {\bm{\phi}} \in \Phi,\; 1 \leq i \leq N;
              \]
              \STATE Generate $N$-step predictions from the history:
              \[
              \left(\vec{Y}_{t+1|t}, \ldots, \vec{Y}_{t+N|t}\right) = \hat{f}(\vec{h}_t);
              \]
              \STATE Solve the ECP-MPC problem~\eqref{eq:ecp-mpc} by evaluating the costs of feasible ${\bm{\phi}}$ in parallel;
              \STATE Apply $\vec{u}_t = \vec{u}^\star_{t|t}$;
	\ENDFOR
\end{algorithmic}
\end{algorithm}

For practical implementation of~\eqref{eq:ecp-mpc-naive}, we instead associate $\alpha_t^i$ with each control input sequence. Specifically, we assume that the space of control inputs is finite and denoted by 
\[
\mathcal{U} \coloneqq \{\vec{u}^1, \ldots, \vec{u}^{n_u}\}.
\]
We then introduce the set of multi-indices 
\[
\Phi \coloneqq [n_u]^N = \Bigl\{\langle \phi_0, \ldots, \phi_{N-1} \rangle : 1 \leq \phi_0, \ldots, \phi_{N-1} \leq n_u \Bigr\}
\]
to label the possible control input sequences.
Intuitively, a sequence ${\bm{\phi}} = \langle \phi_0, \ldots, \phi_{N-1}\rangle$ represents a plan to execute $\vec{u}^{\phi_0}, \ldots, \vec{u}^{\phi_{N-1}}$ starting from the current vehicle state.\footnote{In practice, we avoid the computational burden of exhaustively exploring $\Phi$ by selecting $D \ll N$ decision epochs within the prediction horizon $\{1, \ldots, N\}$ and choosing a control input only at each decision epoch, which lasts for $\ell = N/D$ timesteps. This reduces $|\Phi|$ from $n_u^N$ to $n_u^D$.} To be rigorous, we denote an open-loop control plan by 
\[
\vec{u}^{\bm{\phi}} \coloneqq \langle \vec{u}^{\phi_0}, \ldots, \vec{u}^{\phi_{N-1}} \rangle,
\]
and define the corresponding state sequence 
\[
\vec{x}^{\bm{\phi}}_t \coloneqq \langle \vec{x}^{\bm{\phi}}_{t|t}, \ldots, \vec{x}^{\bm{\phi}}_{t+N|t} \rangle
\]
inductively by setting $\vec{x}^{\bm{\phi}}_{t|t} = \vec{x}_t$ and 
\[
\vec{x}^{\bm{\phi}}_{t+i|t} = f_r\big(\vec{x}^{\bm{\phi}}_{t+i-1|t}, \vec{u}^{\phi_i}\big) \quad \text{for } 0 \leq i < N.
\]
Thus, the MPC problem~\eqref{eq:ecp-mpc-naive} can be cast into the following discrete optimization problem:
\begin{align*}
\min_{{\bm{\phi}} \in \Phi} \quad & \sum_{i=0}^{N-1} \ell\big(\vec{x}^{\bm{\phi}}_{t+i|t}, \vec{u}^{\bm{\phi}}_{t+i|t}\big) + \ell_f\big(\vec{x}^{\bm{\phi}}_{t+N|t}\big) \\
\text{s.t.} \quad & \vec{x}^{\bm{\phi}}_{t+i|t} \in \mathcal{X}, \quad 0 \leq i \leq N, \\
& d\big(\vec{x}^{\bm{\phi}}_{t+i|t}, \vec{Y}_{t+i|t}\big) \geq r_{\text{safe}} + \mathsf{R}^{\vec{x}^{\bm{\phi}}_{t+i|t}}_{t+i|t}, \quad 0 \leq i \leq N.
\end{align*}
Computing $\mathsf{R}^{\vec{x}^{\bm{\phi}}_{t+i|t}}_{t+i|t}$ requires the associated ACP parameter $\alpha^i_t\big(\vec{x}^{\bm{\phi}}_{t+i|t}\big)$ to be continuously updated from the beginning. Unfortunately, since $\vec{x}^{\bm{\phi}}_{t+i|t}$ depends on $\vec{x}_t$, it is challenging to determine in advance the set of future states and update their ACP parameters. Therefore, in contrast to~\eqref{eq:ecp-mpc-naive}, we adopt a different calibration procedure that results in an interval size $\mathsf{R}^{\bm{\phi}}_{t+i|t}$ distinct from $\mathsf{R}^{\vec{x}^{\bm{\phi}}_{t+i|t}}_{t+i|t}$.

For each $1 \leq i \leq N$ and each ${\bm{\phi}} = \langle\phi_0, \ldots, \phi_{N-1}\rangle \in \Phi$, we evaluate the score function associated with $\vec{x}^{\bm{\phi}}_{t+i|t}$ on the calibration set $\mathscr{D}_{t+i|t}$:
\begin{align}\label{eq:score-data}
\mathscr{S}^{\bm{\phi}}_{t+i|t} \coloneqq \Bigl\{ s_i^{\vec{x}^{\bm{\phi}}_{t+i|t}}(\vec{h}_{t'-i}, \vec{Y}_{t'}) \Bigr\}_{t' = t - M + 1}^{t}.
\end{align}
Then, the empirical quantile of the most recent $M$ scores is estimated as
\begin{equation}\label{eq:egocentric-quantile}
\mathsf{R}^{\bm{\phi}}_{t+i|t} \coloneqq \mathsf{Quantile}_{1 - \alpha^{\bm{\phi}}_{t+i|t}} \Bigl( \mathscr{S}^{\bm{\phi}}_{t+i|t} \Bigr),
\end{equation}
where the ACP parameters $\alpha_{t+i|t}^{\bm{\phi}}$ are updated for all ${\bm{\phi}} \in \Phi$ and $1 \leq i \leq N$ according to the rule
\begin{equation}\label{eq:ecp-alpha}
\alpha^{\bm{\phi}}_{t+i+1|t+1} = \alpha^{\bm{\phi}}_{t+i|t} + \gamma \Bigl( \alpha - \mathbb{I}\Bigl[ \vec{Y}_{t} \notin \mathsf{C}^{\bm{\phi}}_{t|t-i} \Bigr] \Bigr),
\end{equation}
with the confidence set around $\vec{Y}_{t+i}$ defined as
\[
\mathsf{C}^{\bm{\phi}}_{t+i|t} \coloneqq \Bigl\{ \vec{Y} \in \mathcal{Y} : s^{\vec{x}^{\bm{\phi}}_{t+i|t}}_i (\vec{h}_t, \vec{Y}) \leq \mathsf{R}^{\bm{\phi}}_{t+i|t} \Bigr\}.
\]
Consequently, rather than running ACP for all $\vec{x} \in \mathcal{X}$, we restrict ACP to a manageable subset corresponding to each ${\bm{\phi}} \in \Phi$. An important observation is that all of $\mathscr{S}^{\bm{\phi}}_{t+i|t}$, $\mathsf{R}^{\bm{\phi}}_{t+i|t}$, and $\mathsf{C}^{\bm{\phi}}_{t+i|t}$ depend only on the initial $i$-tuple $\langle \phi_0, \ldots, \phi_{i-1} \rangle$ of ${\bm{\phi}}$, since $\vec{x}^{\bm{\phi}}_{t+i|t}$ is determined by applying $\vec{u}^{\phi_0}, \ldots, \vec{u}^{\phi_{i-1}}$ to $\vec{x}_t$. Therefore, when $i=1$, there are only $n_u$ different confidence sets, which we denote by $\mathsf{C}^{1}_{t+1|t}, \ldots, \mathsf{C}^{n_u}_{t+1|t}$.

\noindent
\textbf{\textit{Remark.}} The described online calibration procedure must now adapt to additional drifts caused by changes in the vehicle state, since $\vec{x}^{\bm{\phi}}_{t+i|t}$ depends on both ${\bm{\phi}}$ and $\vec{x}_t$. This may introduce extra conservativeness, as the proposed confidence sets might become invalid very rapidly. Thus, we believe that the computational benefits of the suggested scheme come at the cost of adaptation speed, which directly affects the conservativeness of the method.

Analogous to~\eqref{eq:original-cp-asymptotic}, we can establish an asymptotic coverage guarantee for $\mathsf{C}^{\bm{\phi}}_{t+i|t}$.

\begin{proposition}\label{prop:asymptotic-coverage}
For all ${\bm{\phi}} \in \Phi$, $1 \leq i \leq N$, and $\alpha \in (0, 1)$, the following holds:
\begin{equation}\label{eq:asymptotic-coverage}
\lim_{T \to \infty} \frac{1}{T}\sum_{t = 0}^{T-1}\mathbb{I}\Bigl[ \vec{Y}_{t+i} \in \mathsf{C}^{\bm{\phi}}_{t+i|t} \Bigr] = 1 - \alpha, \quad \text{w.p.}\; 1.
\end{equation}
In particular, for $i = 1$, the following holds for each $1 \leq \phi \leq n_u$ with probability 1:
\begin{equation}\label{eq:asymptotic-coverage-next}
\lim_{T \to \infty} \frac{1}{T}\sum_{t = 0}^{T-1}\mathbb{I}\Bigl[ \vec{Y}_{t+1} \in \mathsf{C}^{\bm{\phi}}_{t+1|t} \Bigr] = 1 - \alpha.
\end{equation}
\end{proposition}

\begin{proof}
The result in~\eqref{eq:asymptotic-coverage} follows similarly to~\cite[Proposition 4.1]{gibbs2021adaptive}, with the additional consideration of delayed observations. The key is to show that 
\begin{equation}\label{eq:alpha-bound}
    \alpha_{t+i|t}^{\bm{\phi}} \in [-(i+1)\gamma, 1+(i+1)\gamma],
\end{equation}
for all $1 \leq i \leq N$, ${\bm{\phi}} \in \Phi$, and $t \geq 0$, which is independent of $t$. Note that this bound differs from~\cite[Lemma 4.1]{gibbs2021adaptive} by $i\gamma$, since we must wait an additional $i$ steps to determine whether the confidence set $\mathsf{C}^{\bm{\phi}}_{t+1|t}$ constructed at time $t$ actually covers $\vec{Y}_{t+i}$. The detailed proof of this bound is provided in Appendix~\ref{app:proof-prop2}.
\end{proof}

We now finalize our MPC formulation, termed {Egocentric CP-MPC} (or {ECP-MPC}), as follows:
\begin{align}\label{eq:ecp-mpc}
\begin{split}
\min_{{\bm{\phi}} \in \Phi} \quad & \sum_{i=0}^{N-1} \ell\big(\vec{x}^{\bm{\phi}}_{t+i|t}, \vec{u}^{\bm{\phi}}_{t+i|t}\big) + \ell_N\big(\vec{x}^{\bm{\phi}}_{t+N|t}\big) \\
\text{s.t.} \quad & \vec{x}^{\bm{\phi}}_{t+i|t} \in \mathcal{X}, \quad 0 \leq i \leq N, \\
& d\big(\vec{x}^{\bm{\phi}}_{t+i|t}, \vec{Y}_{t+i}\big) \geq r_{\text{safe}} + \mathsf{R}^{\bm{\phi}}_{t+i|t}, \quad 0 \leq i \leq N.
\end{split}
\end{align}
An outline of the MPC, together with the underlying calibration procedure, is summarized in Algorithm~\ref{alg:ecp-mpc}. Notably, the safety constraints in~\eqref{eq:ecp-mpc} are free from any discretization error, which is attributed to our choice of discretizing the input space rather than the state space.

Assuming the feasibility of~\eqref{eq:ecp-mpc}, we obtain the following safety guarantee.

\begin{theorem}\label{thm:safety}
Suppose that the ECP-MPC formulation~\eqref{eq:ecp-mpc} is feasible for all $t \geq 0$, and denote the optimal solution at time $t$ by 
\[
{\bm{\phi}}^\star_t \coloneqq \bigl(\phi_0^\star(t), \ldots, \phi_{N-1}^\star(t)\bigr),
\]
with corresponding state and control sequences 
\[
\vec{x}^\star_t \coloneqq \bigl(\vec{x}^\star_{t|t}, \ldots, \vec{x}^\star_{t+N|t}\bigr) \quad \text{and} \quad \vec{u}^\star_t \coloneqq \bigl(\vec{u}^\star_{t|t}, \ldots, \vec{u}^\star_{t+N-1|t}\bigr).
\]
Consider the closed-loop system $\{\vec{x}_t\}$ obtained by applying the first control input of the MPC solution, i.e., $\vec{u}_t = \vec{u}^\star_{t|t}$. Then, the system is asymptotically $(1 - n_u\alpha)$-safe, meaning that
\[
\liminf_{t\to\infty} \frac{1}{T}\sum_{t=0}^{T-1} \mathbb{I}\Bigl[ \vec{x}_t \in \mathcal{S}_t \Bigr] \geq 1 - n_u\alpha, \quad \text{w.p.}\; 1.
\]
\end{theorem}

\begin{proof}
From the feasibility of~\eqref{eq:ecp-mpc}, we have
\begin{equation}\label{eq:next-state-constraint}
d\Bigl(\vec{x}^{\star}_{t+1|t}, \vec{Y}_{t+1|t}\Bigr) - \mathsf{R}^{\phi^\star_1(t)}_{t+1|t} \geq r_{\text{safe}}.
\end{equation}
Combined with the condition $\vec{Y}_{t+1} \in \mathsf{C}^{\phi^\star_1(t)}_{t+1|t}$, the inequality~\eqref{eq:next-state-constraint} implies that $\vec{x}^{\star}_{t+1|t} \in \mathcal{S}_{t+1}$. By the definition of $\mathsf{C}^{\phi^\star_1(t)}_{t+1|t}$, the condition $\vec{Y}_{t+1} \in \mathsf{C}^{\phi^\star_1(t)}_{t+1|t}$ ensures that
\[
d\Bigl(\vec{x}^{\star}_{t+1|t}, \vec{Y}_{t+1}\Bigr) \geq d\Bigl(\vec{x}^{\star}_{t+1|t}, \vec{Y}_{t+1|t}\Bigr) - \mathsf{R}^{\phi^\star_1(t)}_{t+1|t},
\]
which is bounded below by $r_{\text{safe}}$ due to feasibility. Therefore, we obtain
\begin{align}
\mathbb{I}\Bigl[ \vec{x}^{\star}_{t+1|t} \in \mathcal{S}_{t+1} \Bigr] &\geq \mathbb{I}\Bigl[ \vec{Y}_{t+1} \in \mathsf{C}^{\phi^\star_1(t)}_{t+1|t} \Bigr] \notag\\[1mm]
&\geq \mathbb{I}\Bigl[ \vec{Y}_{t+1} \in \mathsf{C}^{\phi}_{t+1|t}\;\text{for all } 1 \leq \phi \leq n_u \Bigr] \notag\\[1mm]
&\geq 1 - \sum_{\phi=1}^{n_u} \mathbb{I}\Bigl[ \vec{Y}_{t+1} \notin \mathsf{C}^{\phi}_{t+1|t} \Bigr], \label{eq:union-bound}
\end{align}
where the last inequality follows from the union bound. Consequently,
\begin{align*}
\liminf_{T\to\infty} \frac{1}{T}\sum_{t=0}^{T-1} \mathbb{I}\Bigl[ \vec{x}^{\star}_{t+1|t} \in \mathcal{S}_{t+1} \Bigr] 
&\geq 1 - \limsup_{T\to\infty} \frac{1}{T}\sum_{t=0}^{T-1} \sum_{\phi=1}^{n_u} \mathbb{I}\Bigl[ \vec{Y}_{t+1} \notin \mathsf{C}^{\phi}_{t+1|t} \Bigr] \\
&\geq 1 - \sum_{\phi=1}^{n_u} \limsup_{T\to\infty} \frac{1}{T}\sum_{t=0}^{T-1} \mathbb{I}\Bigl[ \vec{Y}_{t+1} \notin \mathsf{C}^{\phi}_{t+1|t} \Bigr] \\
&= 1 - n_u\alpha,
\end{align*}
where the last equality follows from~\eqref{eq:asymptotic-coverage-next}. Since $\vec{x}_{t+1} = \vec{x}^{\star}_{t+1|t}$, the desired safety guarantee holds.
\end{proof}

\noindent
\textbf{\textit{Remark.}} The constant $n_u$ arises from the union bound in~\eqref{eq:union-bound}. This bound is tight when the events $\{\vec{Y}_{t+i} \in \mathsf{C}^{\phi}_{t+i|t}\}_{\phi=1}^{n_u}$ are disjoint. To eliminate this constant, one would need to precisely quantify the covariance of $\{\mathscr{S}^{\phi}_{t+1|t}\}_{\phi=1}^{n_u}$ in~\eqref{eq:score-data}, which is generally challenging. However, we believe that these values are strongly positively correlated since the corresponding $\vec{x}^{\bm{\phi}}_{t+1|t}$ are similar, ensuring that the confidence sets $\mathsf{C}^{\phi}_{t+1|t}$ are close. In practice, the observed asymptotic safety rates are significantly higher than $1 - n_u\alpha$.

The optimization problem~\eqref{eq:ecp-mpc} can be solved efficiently by leveraging parallelization, as the feasibility and cost of each candidate solution can be evaluated concurrently. In our experiments, a single instance of~\eqref{eq:ecp-mpc} runs in under 0.05 sec on an Intel i9-10940X CPU.

\section{Experimental Results}\label{sec:exp-res}

\begin{table}[t!]
\centering
\caption{Evaluation Metrics across Scenarios and Controllers.}
\vspace{0.1in}
\begin{tabular}{ll|p{0.75cm}p{1.2cm}p{1.0cm}p{0.75cm}}
\toprule
\multicolumn{2}{c|}{} & \multicolumn{4}{c}{Performance Metric} \\[0.5ex]
\midrule
Scenario & Controller & \textit{Collis.} & \textit{Cost} & \textit{Trav.} & \textit{Infeas.} \\
\midrule
\multirow{4}{*}{\texttt{zara1}} 
      & ACP-MPC   &        0.042 &  588.98 &        91.33 &        0.491  \\
      & CC        &        \textbf{0.012} &  921.32 &        91.00 &         $-$  \\
      & ECP-MPC   &        0.034 &  \textbf{506.81} &        \textbf{64.67} &        \textbf{0.185}  \\
\midrule
\multirow{4}{*}{\texttt{zara2}} 
      & ACP-MPC   &        0.029 & 2143.52 &        \textbf{93.33} &        0.672  \\
      & CC        &        \textbf{0.005} & 2036.61 &        97.33 &         $-$  \\
      & ECP-MPC   &        0.016 & \textbf{1099.20} &        96.33 &        \textbf{0.371}  \\
\midrule
\multirow{4}{*}{\texttt{univ}} 
      & ACP-MPC   &        \textbf{0.000} & 2299.62 &       300.00 &        \textbf{0.344}  \\
      & CC        &        0.007 & 1494.38 &       300.00 &         $-$  \\
      & ECP-MPC   &        0.093 &  \textbf{448.87} &       \textbf{262.00} &        0.684 \\
\midrule
\multirow{4}{*}{\texttt{hotel}} 
      & ACP-MPC   &        0.020 &  172.86 &        81.00 &        0.505         \\
      & CC        &        \textbf{0.000} &  193.42 &        99.00 &         $-$        \\
      & ECP-MPC   &        0.005 &  \textbf{139.96} &  \textbf{68.67} &        \textbf{0.467}       \\
\midrule
\multirow{4}{*}{\texttt{eth}} 
      & ACP-MPC   &        0.017 &  726.16 &        \textbf{79.00} &        \textbf{0.320}         \\
      & CC        &        \textbf{0.006} &  845.48 &        99.00 &         $-$        \\
      & ECP-MPC   &        0.012 &  \textbf{690.76} &  100 &        0.737       \\
\bottomrule
\end{tabular}
\label{tab:metric}
\end{table}

To evaluate the effectiveness of ECP-MPC, we consider dynamic navigation scenarios in which an ego-vehicle must reach specified goals while avoiding multiple pedestrians. We assume that the state of the ego-vehicle is given by its 2D position and orientation, i.e., $\vec{x} = [x, y, \theta]^\top$, and that the control inputs are $\vec{u} = [v, \omega]^\top$, representing the linear and angular velocities. The vehicle follows a unicycle model:
\[
f_r(\vec{x}, \vec{u}) = \vec{x} + h \begin{bmatrix} v\cos\theta \\ v\sin\theta \\ \omega \end{bmatrix},
\]
with a time step $h = 0.4$. We discretize the control inputs to obtain 
\[
\mathcal{U} = \{-0.8,\, 0,\, 0.8\} \times \{-0.7,\, 0,\, 0.7\},
\]
so that $n_u = 9$. Given a goal position $[x_{\text{goal}}, y_{\text{goal}}]^\top$, we consider the cost functions
\[
\ell(\vec{x}, \vec{u}) = (x - x_{\text{goal}})^2 + (y - y_{\text{goal}})^2 + \vec{u}^\top \vec{R} \vec{u} \quad \text{and} \quad \ell_f(\vec{x}) = 10(x - x_{\text{goal}})^2 + 10(y - y_{\text{goal}})^2,
\]
with $\vec{R} = 10^{-3}\vec{I}_2$.

To simulate the motions of dynamic obstacles, we use the ETH-UCY dataset~\cite{lerner2007crowds,pellegrini2009you}, which provides 2.5 Hz positional annotations of pedestrians in five scenarios: \texttt{zara1}, \texttt{zara2}, \texttt{univ}, \texttt{hotel}, and \texttt{eth}. As the prediction model, we use Trajectron++~\cite{salzmann2020trajectron++}, a state-of-the-art model that efficiently captures both motion history and interactions among pedestrians. The model takes histories of length $H = 8$ and outputs predictions for $N = 12$ steps. We evaluate ACP-MPC~\cite{dixit2023adaptive} and the Conformal Controller (CC)~\cite{lekeufack2024conformal} as baselines. We set $\alpha = 0.1$ for both ACP-MPC and ECP-MPC, while the desired risk level of the conformal controller is chosen as $\varepsilon = -2$.

For the \texttt{zara1}, \texttt{zara2}, \texttt{hotel}, and \texttt{eth} scenarios, we run three episodes, each consisting of $t_{\max} = 100$ steps (approximately 40 seconds). For \texttt{univ}, we run a single episode with $t_{\max} = 300$ steps due to the complexity of the environment.\footnote{The source code used for the experiments is available at \url{https://github.com/CORE-SNU/ECP-MPC}.}

All methods are evaluated using the following performance metrics:
\begin{itemize}
\item \textbf{Travel Time (\textit{Trav.})} --- The number of time steps required to reach the specified goal, denoted by $\tau$. If the goal is not reached by $t_{\max}$, then $\tau = t_{\max}$.
\item \textbf{Collision Rate (\textit{Collis.})} --- The fraction of time steps during which $\vec{x}_t$ is outside the safe set $\mathcal{S}_t$, computed as $\sum_{t=0}^{\tau-1}\mathbb{I}[\vec{x}_t \notin \mathcal{S}_t] / \tau$.
\item \textbf{Cost (\textit{Cost})} --- The average cost incurred, given by 
\[
\frac{1}{\tau}\sum_{t=0}^{\tau - 1}\Biggl(\sum_{i=0}^{N-1} \ell(\vec{x}_{t+i|t}, \vec{u}_{t+i|t}) + \ell_N(\vec{x}_{t+N|t})\Biggr).
\]
\item \textbf{Infeasibility Rate (\textit{Infeas.})} --- The fraction of time steps at which the MPC problem becomes infeasible.
\end{itemize}

The evaluation results are summarized in Table~\ref{tab:metric}. Across all scenarios, ECP-MPC achieves the lowest cost, confirming its cost-efficiency as noted in Proposition~\ref{prop:less-conservative}. Although the conformal controller generally yields the lowest collision rates (except in \texttt{univ}), ECP-MPC outperforms ACP-MPC with slightly lower collision rates—likely because ACP-MPC frequently becomes infeasible and freezes, which can lead to collisions. For both ACP-MPC and ECP-MPC, the collision rates remain well below $\alpha = 0.1$ (and notably below $n_u\alpha$, as indicated in Theorem~\ref{thm:safety}).

\begin{figure}[t]
    \centering
    \includegraphics[width=0.75\linewidth]{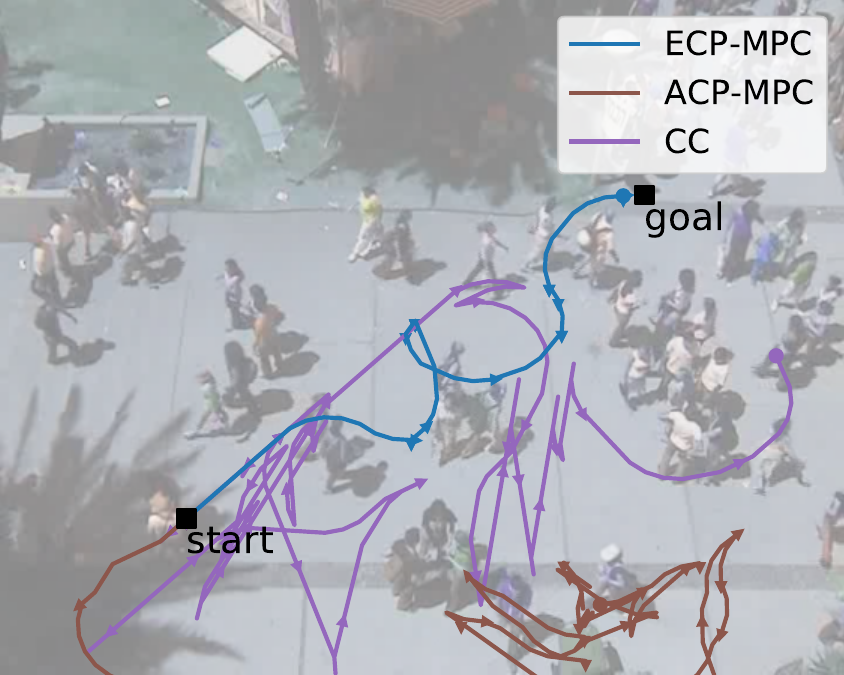}
    \caption{Trajectories of the vehicle produced by ECP-MPC, ACP-MPC, and the conformal controller in the \texttt{univ} scenario.}
    \label{fig:univ-example}
\end{figure}

Figure~\ref{fig:univ-example} shows the ego-vehicle trajectories produced by each method in the \texttt{univ} scenario, the most challenging environment due to its high pedestrian density. Although ECP-MPC is the only method that reaches the goal within $t_{\max}$, it experiences a high infeasibility rate because the ego-vehicle drives toward the center of the scene and becomes temporarily trapped by large crowds approaching from all directions. In contrast, ACP-MPC exhibits a lower infeasibility rate but tends to wander along the scene boundary, which is farther from the goal yet less crowded. Meanwhile, the conformal controller brings the vehicle near the goal but requires substantial back-and-forth motion due to oscillatory behavior.

Finally, in the \texttt{eth} scenario, ACP-MPC outperforms ECP-MPC. In this scenario, many pedestrians are tracked near the center of the scene, causing the egocentric score function to spike unexpectedly when some obstacles suddenly appear close to the vehicle. In such cases, the obstacle-centric score function is preferable because its values are less susceptible to these sudden spikes. Nevertheless, we believe that such situations are rare in real-world applications, where dynamic obstacles typically enter and exit near the boundary of the vehicle’s sensing range.

\section{Conclusions and Future Work}
We introduce the notion of an egocentric score function, which forms the basis of a novel MPC scheme that is significantly more cost-efficient than the existing ACP-MPC. The method ensures asymptotic safety, and empirical evidence demonstrates its effectiveness compared to established baselines.

Although our approach is presented in the context of navigation scenarios where only collision-related safety is considered, we believe it can be generalized to a broader class of optimal control problems with more comprehensive safety constraints. Another interesting direction involves scenarios with imperfect perception. In such cases, the system state must be inferred from noisy, partial observations, and the method must also account for potential perception errors.

\section*{Acknowledgment}
The authors would like to thank Prof. Ram Vasudevan at the University of Michigan for his insightful discussion.

\appendix

\section{ECP-MPC with State Space Discretization}\label{app:state-discretization}
In this section, we provide a rough sketch of~\eqref{eq:ecp-mpc-naive} that relies on state space discretization. First, we introduce a discretization 
\[
\bar{\mathcal{X}} = \{\vec{x}^1, \ldots, \vec{x}^{n_x} \}
\]
of the state space, with resolution specified by a parameter $\delta > 0$, such that
\[
\mathcal{X} \subseteq \bigcup_{p=1}^{n_x} B(\vec{x}^p, \delta) = \bigcup_{p=1}^{n_x} \{\vec{x}\in \mathcal{X} : \|\vec{x} - \vec{x}^{p}\| \leq \delta \}.
\]
Instead of maintaining $\alpha_t^i(\vec{x})$ for all $\vec{x} \in \mathcal{X}$ (which is impractical), we compute finitely many values $\alpha_t^i(\vec{x}^p)$ for $1 \leq p \leq n_x$. Then, for each discretized state $\vec{x}^p$, we check the following constraint, augmented with the resolution parameter $\delta$:
\[
\mathcal{P}_{t+i|t} \coloneqq \Bigl\{ p : d\big(\vec{x}^p, \vec{Y}_{t+i|t}\big) \geq r_{\text{safe}} + \mathsf{R}^{\vec{x}^p}_{t+i|t} \;{\color{violet}+ \delta} \Bigr\}.
\]
Then, the egocentric safe set $\mathcal{S}^{\text{ego}}_{t+i|t}$ admits the following  {approximation}:
\[
\bar{\mathcal{S}}^{\text{ego}}_{t+i|t} \coloneqq \bigcup_{p \in \mathcal{P}_{t+i|t}} B(\vec{x}^p, \delta),
\]
whose validity follows readily from the triangle inequality:
$d(\vec{x}, \vec{Y}_{t+i}) \geq d(\vec{x}^p, \vec{Y}_{t+i}) - \|\vec{x} - \vec{x}^p\|$.

\section{Proof of Proposition~\ref{prop:asymptotic-coverage}}\label{app:proof-prop2}
\begin{proof}
In this proof, we show that 
\begin{equation}\label{eq:alpha-bound}
    \alpha_{t+i|t}^{\bm{\phi}} \in [-(i+1)\gamma,\, 1+(i+1)\gamma],
\end{equation}
for all $1\leq i \leq N$, $\bm{\phi}\in \Phi$, and $t \geq 0$, where the bound is independent of $t$. Once this bound is established, the remainder of the proof follows by unrolling and rearranging~\eqref{eq:ecp-alpha}:
\begin{align*}
\frac{1}{T}\sum_{t=0}^{T-1}\mathbb{I}\Bigl[ \vec{Y}_{\tau} \in \mathsf{C}^{\bm{\phi}}_{t+i|t} \Bigr] 
&= (1-\alpha) + \frac{\alpha^{\bm{\phi}}_{T+2i|T+i} - \alpha^{\bm{\phi}}_{2i|i}}{\gamma T} \\
&= (1-\alpha) + O(T^{-1}).
\end{align*}

We now prove the lower bound in~\eqref{eq:alpha-bound}. For simplicity, fix $i$ and $\bm{\phi}$, and define 
\[
\beta_t \coloneqq \alpha^{\bm{\phi}}_{t+i|t}.
\]
Partition the set $\{t : \beta_t \leq 0 \}$ into disjoint intervals $[\ell_j, u_j)$ with $u_j \in \mathbb{N} \cup \{\infty\}$ and $u_j < \ell_{j+1}$. It suffices to show that $\beta_t \geq -(i+1)\gamma$ on each such interval. 

To that end, observe that by construction, we have $\beta_{\ell_j - 1} > 0$. Thus, for $t \in [\ell_j, u_j)$, it holds that
\[
\beta_t \geq \beta_{\ell_j - 1} - (t-\ell_j+1)\gamma (1-\alpha).
\]
If $u_j-\ell_j \leq i$, then for all $t \in [\ell_j, u_j)$ we have 
\[
-(t-\ell_j+1)\gamma (1-\alpha) \geq -i\gamma(1-\alpha),
\]
which in turn implies $\beta_t \geq -i\gamma(1-\alpha) \geq -(i+1)\gamma$, since $1-\alpha \leq 1$. 

If, on the other hand, $u_j-\ell_j > i$, we can divide the interval into two parts: $[\ell_j, \ell_j+ i)$ and $[\ell_j+i, u_j)$. Notice that whenever $\beta_t \leq 0$, we have $\mathsf{R}^{\bm{\phi}}_{t+i|t} = \infty$, which forces 
\[
\mathbb{I}\Bigl[\vec{Y}_{t+i} \in \mathsf{C}^{\bm{\phi}}_{t+i|t}\Bigr] = 1,
\]
and consequently, the update rule yields
\[
\beta_{t+i+1} = \beta_{t+i} + \gamma \alpha.
\]
Thus, on $[\ell_j+i, u_j)$ the sequence $\beta_t$ is increasing. It follows that
\[
\min_{\ell_j\leq t < u_j} \beta_t = \min_{\ell_j \leq t \leq \ell_j+i} \beta_t \geq -(i+1)\gamma(1-\alpha).
\]
Since $-(i+1)\gamma(1-\alpha) \geq -(i+1)\gamma$, we obtain 
\[
\min_{t\in [\ell_j, u_j]} \beta_t > -(i+1)\gamma,
\]
which proves the lower bound in~\eqref{eq:alpha-bound}. The upper bound can be shown using a similar argument.
\end{proof}

\bibliographystyle{IEEEtran}
\bibliography{reference}

\end{document}